\documentclass{article}

\usepackage{arxiv}

\usepackage[authoryear,longnamesfirst]{natbib}
\usepackage[utf8]{inputenc} % allow utf-8 input
\usepackage[T1]{fontenc}    % use 8-bit T1 fonts
\usepackage{hyperref}       % hyperlinks
\usepackage{url}            % simple URL typesetting
\usepackage{booktabs}       % professional-quality tables
\usepackage{amsfonts}       % blackboard math symbols
\usepackage{nicefrac}       % compact symbols for 1/2, etc.
\usepackage{microtype}      % microtypography
\usepackage{lipsum}
\usepackage{graphicx}
\usepackage{booktabs}
\usepackage{amsmath}
\usepackage{algorithm,algpseudocode}
\usepackage{makecell}
\usepackage{boldline, makecell}
\usepackage{multirow}
\usepackage{subcaption}
\usepackage{mathtools}
\usepackage{float}

\newcommand\blfootnote[1]{%
  \begingroup
  \renewcommand\thefootnote{}\footnote{#1}%
  \addtocounter{footnote}{-1}%
  \endgroup
}

\graphicspath{ {./images/} }

\DeclareUnicodeCharacter{2212}{-}

\title{Emotion-guided Cross-domain Fake News Detection using Adversarial Domain Adaptation}

\author{
 Arjun Choudhry\footnotemark[1] \\
  Biometric Research Laboratory\\
  Delhi Technological University\\
  New Delhi, India\\
  \texttt{choudhry.arjun@gmail.com} \\
  \And
 Inder Khatri\footnotemark[1] \\
  Biometric Research Laboratory\\
  Delhi Technological University\\
  New Delhi, India\\
  \texttt{inderkhatri999@gmail.com} \\
  \And
 Arkajyoti Chakraborty \\
  Biometric Research Laboratory\\
  Delhi Technological University\\
  New Delhi, India\\
  \texttt{arkajyotichakraborty\_2k19ep022@dtu.ac.in} \\
  \And
  Dinesh Kumar Vishwakarma \\
  Biometric Research Laboratory\\
  Delhi Technological University\\
  New Delhi, India\\
  \texttt{dinesh@dtu.ac.in} \\
  \And
  Mukesh Prasad \\
  School of Computer Science \\
  University of Technology Sydney\\
  Ultimo, Australia\\
  \texttt{mukesh.prasad@uts.edu.au} \\  
}

\begin{document}
\maketitle

\begin{abstract}
Recent works on fake news detection have shown the efficacy of using emotions as a feature or emotions-based features for improved performance.  However, the impact of these emotion-guided features for fake news detection in cross-domain settings, where we face the problem of domain shift, is still largely unexplored. In this work, we evaluate the impact of emotion-guided features for cross-domain fake news detection, and further propose an emotion-guided, domain-adaptive approach using adversarial learning. We prove the efficacy of emotion-guided models in cross-domain settings for various combinations of source and target datasets from FakeNewsAMT, Celeb, Politifact and Gossipcop datasets.\blfootnote{*Equal Contribution}
\end{abstract}


\section{Introduction}

In recent years, our reliance on social media as a source of information has increased multi-fold, bringing along exponential increase in the spread of \emph{fake news}. To counter this, researchers have proposed various approaches for fake news detection \citep{Defend, NEP}. However, models trained on one domain are often brittle and vulnerable to incorrect predictions for the samples of another domain \citep{saikh, perez-rosas}. This is primarily due to the shift between the two domains, as depicted in Figure 1(1). To handle this, some domain-adaptive frameworks \citep{BDANN,DAFD,MDSWS} have been proposed which help align the source and target domains in the feature space to ameliorate domain shift across different problems. These frameworks guide the feature extractors to extract domain-invariant features by aligning the source and target domains in the feature space, thus generalizing well across domains. However, due to the absence of labels in the target-domain data, the adaptation is often prone to negative transfer, which can disturb the class-wise distribution and affect the discriminability of the final model, as shown in Figure 1(2).

Some recent studies have observed a correlation between the veracity of a text and its emotions. There exists a prominent affiliation for certain emotions with fake news, and for other emotions with real news \citep{tweets_mit}, as illustrated in Figure 1(3). Further, some works have successfully utilized emotions as features, or emotion-guided features to aid in fake news detection \citep{Dual_1, Dual_2,Emo_FND_AAAI}. However, we observe that these works only consider the in-domain setting for evaluation, without analyzing the robustness of these frameworks to domain shift in cross-domain settings. This is another important direction that needs to be explored.

\begin{figure}[b!]
     \centering
     \includegraphics[width = 0.6\columnwidth]{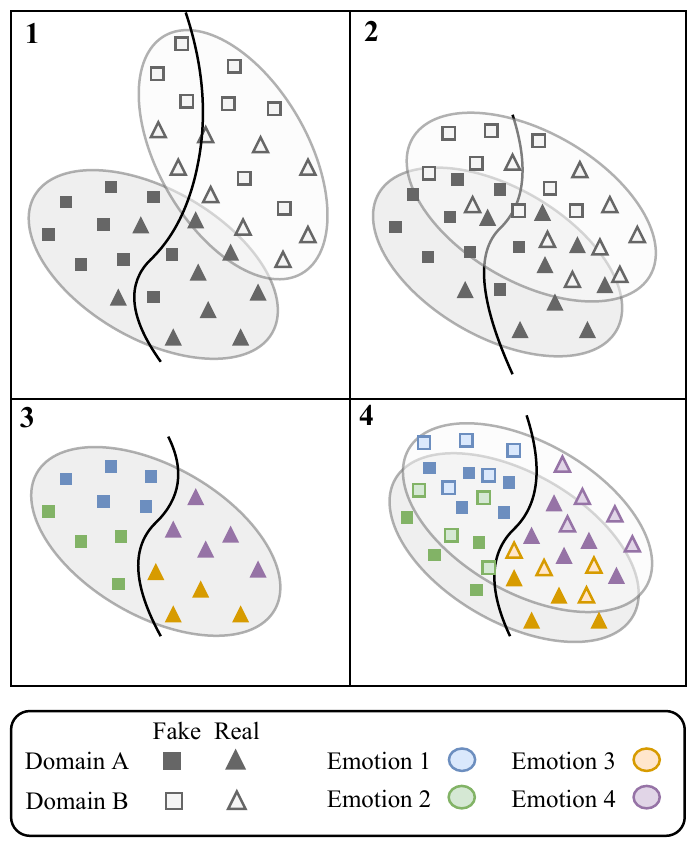}
     \hfill
   \caption{(1) Cross-domain texts not aligned. (2) Domain adaptation leads to some alignment. (3) Emotion-guided classification in one domain. (4) Emotion-guided domain adaptation leads to improved alignment of the two domains.}
   \label{DA_Flowchart} 
\end{figure}

In this paper, we study the efficacy of emotion-aided models in capturing better generalizable features for cross-domain fake news detection. Table \ref{Results} shows the improvements observed in various cross-domain settings when our emotion-guided models were evaluated in cross-domain settings. We observe that emotion-guided frameworks show improved performance in cross-domain settings, as compared to their baseline models without the said emotion-aided features, thus underscoring the generalized feature extraction in emotion-aided models. We further propose an emotion-guided unsupervised domain adaptation framework, which utilizes emotion labels in a multi-task adversarial setting for better feature alignment across domains. The emotion labels for emotion classification, trained parallel to the fake news detection branch in the multi-task learning setup, help provide additional supervision for improved alignment during domain adaptation, mitigating the issue of incorrect alignment of domains. This is illustrated in Figure 1(4)). This leads to better discriminability. We experimentally prove the efficacy of our approach across a variety of datasets in cross-domain settings for various combinations of single-task or multi-task, domain-adaptive or non-adaptive, emotion-guided or unguided settings on the accuracy of the models.

Our contributions can be summarized as follows:
\begin{itemize}
    \item We suggest the use of emotion classification as an auxiliary task for improved fake news detection in cross-domain settings, indicating the applicability of emotion-guided features across domains.
    \item We compare how Ekman's and Plutchik's base emotion classes individually affect the performance of our multi-task domain-adaptive framework, and if there are meaningful differences between them.
    \item We propose an emotion-guided domain-adaptive framework for fake news detection across domains. We show that domain-adaptive fake news detection models better align the two domains with the help of supervised learning using emotion-aided features.
    \item We evaluate our approach on a variety of source and target combinations from four datasets. Our results prove the efficacy of our approach.
\end{itemize}

\section{Related Works}
Several studies over the last few years have explored the correlation of fake news detection with emotions. \citet{Health_Fake_News} \emph{emotionized} text representations using explicit emotion intensity lexicons. \citet{Dual_1} utilized the discrepancies between publisher's emotion and the thread's comments' emotions to detect fake news. However, most of these methods relied upon some additional inputs during evaluation. \citet{Emo_FND_AAAI} proposed an emotion-aided multi-task learning approach, where emotion classification was the auxiliary task implicitly aligning fake news features according to emotion labels. 

Inspired by \citet{DANN}, \citet{BDANN} proposed the first fake news detection work to tackle domain shifts between different datasets. They proposed a multi-modal framework with a Gradient Reversal Layer (GRL) to learn domain-invariant features across different domains and used a joint fake news detector on the extracted features. \citet{DAFD} proposed a robust and generalized fake news detection framework adaptable to a new target domain using adversarial training to make the model robust to outliers and Maximum Mean Difference (MMD)-based loss to align the features of source and target. \citet{MDSWS} extended the problem by treating it as a multi-source domain adaptation task, using the labeled samples from multiple source domains to improve the performance on unlabeled target domains. They also utilized weak labels for weak supervision on target samples to improve performance.

However, no previous work has aligned features between different domains using emotion-guided features and domain adaptation using adversarial training. We show that applying both of these approaches leads to improved performance due to better alignment of inter-domain features.
\section{Proposed Methodology}

\subsection{Datasets, Emotion Annotation \& Preprocessing}
We use the FakeNewsAMT \citep{perez-rosas}, Celeb \citep{perez-rosas}, Politifact\footnote{https://www.politifact.com}, and Gossipcop\footnote{https://www.gossipcop.com} datasets for cross-domain fake news detection. FakeNewsAMT is a multi-domain dataset containing samples from technology, education, business, sports, politics, and entertainment domains. The Celeb dataset has been derived from the web, and contains news about celebrities. Politifact is a web-scrapped dataset containing political news, while Gossipcop contains news extracted from the web, manually annotated via crowd-sourcing and by experts.

\begin{figure*}[t!]
     \centering
     \includegraphics[scale=0.8]{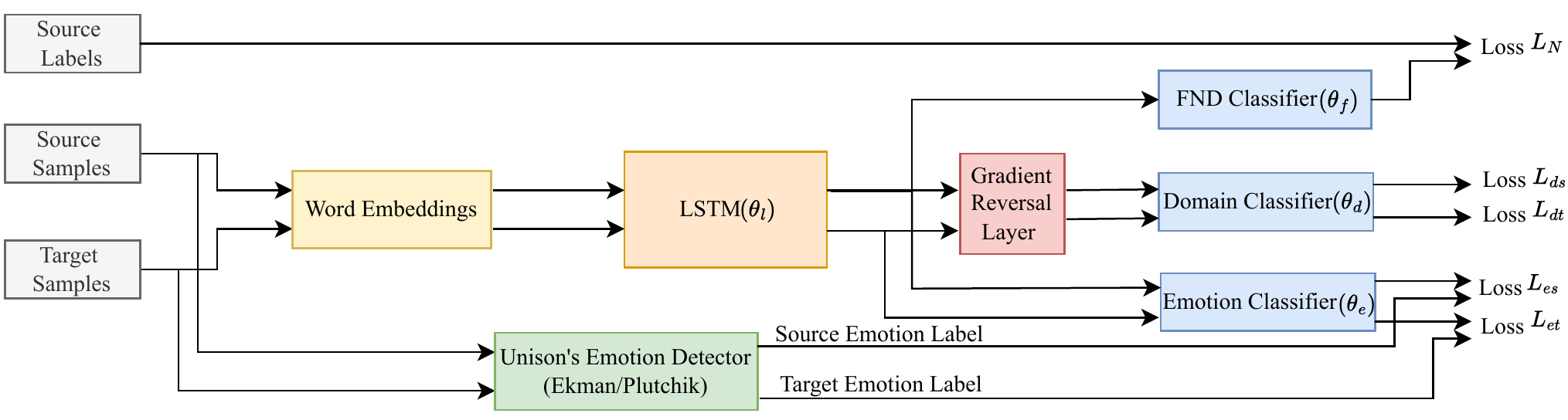}
     \hfill
   \caption{Pictorial depiction of our emotion-guided domain-adaptive approach for cross-domain fake news detection.}
   \label{Flowchart} 
\end{figure*}

We use the Unison model \citep{unison} to annotate all datasets with the core emotions from Ekman's \citep{ekman} (6 emotions: \textit{Joy}, \textit{Surprise}, \textit{Anger}, \textit{Sadness}, \textit{Disgust}, \textit{Fear}) and Plutchik's \citep{plut} (8 emotions: \textit{Joy}, \textit{Surprise}, \textit{Trust}, \textit{Anger}, \textit{Anticipation}, \textit{Sadness}, \textit{Disgust}, \textit{Fear}) emotion theories. During preprocessing, we convert text to lowercase, remove punctuation, and de-contract verb forms (eg. \textit{“I’d”} to \textit{“I would”}).

\subsection{Multi-task Learning}
We use multi-task learning (MTL) to incorporate emotion classification as an auxiliary task to our fake news detection branch. Multi-task learning enables a model to learn the shared features between two or more correlated tasks for improved feature extraction and performance. We use Ekman's or Plutchik's emotions labels for emotion classification branch in our MTL models to see which performs better, and compare the performance with the corresponding single-task (STL) models in domain-adaptive and non-adaptive settings.

\subsection{Emotion-guided Domain-adaptive Framework}

We propose the cumulative use of domain adaptation and emotion-guided feature extraction for cross-domain fake news detection. Our approach aims to improve the feature alignment between different domains using adversarial domain adaptation, by leveraging the correlation between the emotion and the veracity of a text (as shown in Figure \ref{DA_Flowchart}(4)). Figure \ref{Flowchart} shows our proposed framework. We use an LSTM-based \citep{lstm} feature extractor, which is trained using the accumulated loss from fake news classifier, emotion classifier and the discriminator (aids in learning domain-invariant features). LSTM can be replaced with better feature extractors. We used it specifically for easier comparison to non-adapted emotion-guided and non-adapted single-task models. The domain classifier acts as the discriminator. In our proposed framework, we do not use the truth labels for the target dataset for domain adaptation. However, we utilize the target domain emotion labels in our approach to better align the two domains using the emotion labels for supervised learning. The fake news classification loss, emotion classification loss, adversarial loss, and total loss are defined as in Equations 1, 2, 3, and 4:

\begin{equation}
 L_{FND} \ = \ \min\limits_{\theta_{l},\theta_{f}} \sum_{i=1}^{n_{s}} L_{f}^i
\end{equation}

\begin{equation}
 L_{emo} \ = \ \min\limits_{\theta_{l},\theta_{e}} \sum_{i=1}^{n_{s}} L_{es}^i \ + \ \sum_{j=1}^{n_{t}} L_{et}^j))
\end{equation}

\begin{equation}
 L_{adv} \ = \ \min\limits_{\theta_d} (\max\limits_{\theta_l}( \sum_{i=1}^{n_{s}} L_{ds}^i \ + \ \sum_{j=1}^{n_{t}} L_{dt}^j))
\end{equation}

\begin{equation}
 L_{Total} \ = \ (1 - \alpha - \beta) * L_{FND} \ + \ \alpha \ * \ (L_{adv}) \ + \ \beta \ * \ (L_{emo}) 
\end{equation}

where $n_s$ and $n_t$ are number of samples in source and target sets; $\theta_d$, $\theta_f$, $\theta_e$ and $\theta_l$ are parameters for discriminator, fake news classifier, emotion classifier and LSTM feature extractor; $L_{d_s}$ and $L_{d_t}$ are binary crossentropy loss for source and target classification; $L_{es}$ and $L_{et}$ are crossentropy loss for emotion classification; $L_f$ is binary crossentropy loss for Fake News Classifier; $\alpha$ and $\beta$ are weight parameters in $L_{Total}$. We optimised $\alpha$ and $\beta$ for each setting for optimal performance.

We used 300 dimension GloVe \citep{glove} embeddings. All models were trained for up to 50 epochs, stopped when the peak validation accuracy for the in-domain validation set was achieved. We used a batch size of 25 for every experiment. Each model used the Adam optimizer with learning rate 0.0025. We used an LSTM layer with 256 units for feature extraction, while both fake news detection and emotion classification branches consisted of two dense layers each. 

\section{Experimental Analysis \& Results}

We evaluated our proposed approach on various combinations of source and target datasets. Each model was optimized on an in-domain validation set from the source set. Table \ref{Results} illustrates our results proving the efficacy of using emotion-guided models in non-adapted and domain-adapted cross-domain settings. It compares non-adaptive models, domain-adaptive models, and our emotion-guided domain-adaptive models in various settings. MTL(E) and MTL(P) refer to emotion-guided multi-task frameworks using Ekman's and Plutchik's emotions respectively. STL refers to the single-task framework. DA refers to the use of the domain-adaptive framework, containing a discriminator. Some findings observed are:

\begin{table}[t!]
    \centering
    \begin{tabular}{c|c|c|c}
    \hline
    \hline
        \textbf{Source} & \textbf{Target} & \textbf{Setting} & \textbf{Accuracy} \\\hline
        \multirow{6}{*}{FakeNewsAMT} & \multirow{6}{*}{Celeb} & STL & 0.420\\
        & & MTL(E) & 0.520\\
        & & MTL(P) & 0.530\\
        & & DA STL & 0.560\\
        & & DA MTL(E) & 0.540\\
        & & DA MTL(P) & \textbf{0.600}\\
        \hline
        \multirow{6}{*}{Celeb} & \multirow{6}{*}{FakeNewsAMT} & STL & 0.432\\
        & & MTL(E) & 0.471\\
        & & MTL(P) & 0.476\\
        & & DA STL & 0.395\\
        & & DA MTL(E) & 0.501\\
        & & DA MTL(P) & \textbf{0.551}\\
        \hline
        \multirow{6}{*}{Politifact} & \multirow{6}{*}{Gossipcop} & STL & 0.527\\
        & & MTL(E) & 0.555\\
        & & MTL(P) & 0.603\\
        & & DA STL & 0.585\\
        & & DA MTL(E) & \textbf{0.698}\\
        & & DA MTL(P) & 0.671\\
        \hline
        \multirow{6}{*}{Celeb} & \multirow{6}{*}{Gossipcop} & STL & 0.488\\
        & & MTL(E) & 0.501\\
        & & MTL(P) & 0.490\\
        & & DA STL & 0.525\\
        & & DA MTL(E) & 0.555\\
        & & DA MTL(P) & \textbf{0.587}\\
        \hline
        \multirow{6}{*}{FakeNewsAMT} & \multirow{6}{*}{Gossipcop} & STL & 0.451\\
        & & MTL(E) & 0.652\\
        & & MTL(P) & 0.620\\
        & & DA STL & 0.790\\
        & & DA MTL(E) & \textbf{0.805}\\
        & & DA MTL(P) & 0.795\\
        \hline
        \multirow{6}{*}{FakeNewsAMT} & \multirow{6}{*}{Politifact} & STL & 0.363\\
        & & MTL(E) & 0.450\\
        & & MTL(P) & 0.530\\
        & & DA STL & 0.621\\
        & & DA MTL(E) & \textbf{0.704}\\
        & & DA MTL(P) & 0.621\\
    \hline
    \hline
    \end{tabular}
    \caption{Cross-domain evaluation of non-adaptive and adaptive models on FakeNewsAMT, Celeb, Politifact and Gossipcop datasets. Emotion-guided domain-adaptive models (DA MTL(E) and DA MTL(P)) outperform their corresponding STL models in cross-domain settings. Domain-adaptive MTL models consistently outperform baseline STL, non-adaptive MTL and domain-adaptive STL models.}
    \label{Results}
\end{table}

\textbf{MTL(E) and MTL(P) models outperform their STL counterparts in cross-domain settings}, as seen in Table \ref{Results}. This indicates improved extraction of generalizable features by the emotion-guided models, which aids in improved fake news detection across datasets from different domains. MTL(E) and MTL(P) further perform comparably for most settings, and each outperforms the other in three settings respectively.

\textbf{DA STL models generally outperform STL models in cross-domain settings} across multiple combinations of datasets. However, we see the STL model outperformed the DA STL model for Celeb dataset as the source dataset and FakeNewsAMT dataset as target, confirming that unguided adaptation can sometimes lead to negative alignment, reducing the performance of the model.

\textbf{DA MTL(E) and DA MTL(P) models improve performance in cross-domain settings.} Table \ref{Results} shows improved results obtained using the emotion-guided adversarial DA models over their non-adaptive counterparts. This shows the scope for improved feature extraction even after using DA, and emotion-guided models act as a solution aiding in correct alignment of the samples and features extracted by the adaptive framework from different domains. Emotion-guided DA models mitigated the issue of negative alignment when Celeb dataset was the source and FakeNewsAMT dataset the target, where STL model outperformed the DA STL model. The emotion-guided DA models helped correctly align the two domains, leading to significantly improved performance.

\section{Conclusion}
In this work, we showed the efficacy of emotion-guided models for improved cross-domain fake news detection and further presented an emotion-guided domain-adaptive fake news detection approach. We evaluated our proposed framework against baseline STL, emotion-guided MTL, DA STL and emotion-guided DA MTL models for various source and target combinations from four datasets. Our proposed approach led to improved cross-domain fake news detection accuracy, indicating that emotions are generalizable across domains and aid in better alignment of different domains during domain adaptation.

\bibliographystyle{cas-model2-names}

\bibliography{main}

\end{document}